%% file: arxiv.tex
\documentclass{article} 
\usepackage{iclr2027_conference,times}

\input{math_commands.tex}

\usepackage{url}
\usepackage{graphicx}
\usepackage{booktabs}
\usepackage{amsmath,amssymb}
\usepackage{multirow}
\usepackage{placeins}
\usepackage{needspace}
\usepackage{enumitem}
\usepackage{wrapfig}
\usepackage{xcolor}
\usepackage{fancyhdr}

\newcommand{\wc}[1]{w_{c(#1)}}
\newcommand{\xb}{\bar{\boldsymbol x}}
\newcommand{\xbk}[1]{\bar x_{#1}}

\definecolor{cbgreen}{RGB}{34,136,51}
\definecolor{cbblue}{RGB}{0, 119, 187}
\definecolor{cbred}{RGB}{204, 51, 17}
\definecolor{richlilac}{rgb}{0.71, 0.4, 0.82}
\definecolor{mydarkblue}{rgb}{0,0.08,0.45}
\definecolor{circuitred}{RGB}{255, 220, 220}    
\definecolor{circuitblue}{RGB}{210, 225, 245}   
\usepackage[colorlinks=true,citecolor=cbgreen,linkcolor=magenta,urlcolor=mydarkblue]{hyperref}
\usepackage[capitalize,noabbrev]{cleveref}

\title{Who Is the Influencer? \\ Consensus Dynamics in Multi-Agent Systems}

\author{%
  Gouki Minegishi$^{1}$ \quad Jingyuan Feng$^{1}$ \quad Hiroki Furuta \\
  \textbf{Takeshi Kojima}$^{1}$ \quad \textbf{Yusuke Iwasawa}$^{1}$ \quad \textbf{Yutaka Matsuo}$^{1}$ \\
  $^{1}$The University of Tokyo \\
}

\iclrfinalcopy

\begin{document}

\fancypagestyle{plain}{%
  \fancyhf{}%
  \lhead{Preprint}%
  \cfoot{\thepage}%
  \renewcommand{\headrulewidth}{0pt}%
}
\pagestyle{plain}

\maketitle

\begin{abstract}
Multi-agent LLM systems are increasingly used in practice, but their behavior
is complex and hard to understand. We design a simple task to ask which model
has the most influence on the final consensus when several models are mixed in
one system. Experimentally, across seven model families, we find
that the opinions of the other models are absorbed into the opinion of the
stronger model, the one with more parameters. This happens even though the task
requires no pretraining knowledge and has no correct answer. To understand this
phenomenon, we interpret the agents as weighted particles moving on a simplex.
Larger models carry a larger weight and therefore occupy a deeper well in the
energy landscape. A large weight means strong in-context learning: a model with
a large weight amplifies the majority of its context. Larger models are
generally better at in-context learning, and this is what makes the larger
model the influencer. We hope that this work helps to interpret multi-agent
systems, whose behavior is becoming too complex for humans to understand.
\end{abstract}

\section{Introduction}

Language models are increasingly used not as a single model but as a system
of many agents \citep{guo2024masurvey,tran2025masurvey}. Such systems often do
better than one model. When instances of a model debate a question, their
factual accuracy and reasoning improve over a single chain of thought
\citep{du2024debate,liang2024mad,khan2024persuasive}. Performance also
improves as more agents are added \citep{li2024moreagents,qian2025macnet}. Most
recently, OpenAI reported that about ten thousand agents working in parallel had
produced a formally verified, though not yet peer-reviewed, proof of finite-time
blow-up for the Navier--Stokes equations \citep{openai2026navierstokes}. Multi-agent systems also simulate human society
\citep{anthis2025position,beyondstatic2025}: agents live in simulated towns
\citep{park2023generative} and on Moltbook, a social network with no human
members \citep{moltbook2026}, where agents of many kinds interact with one
another.

Multi-agent systems can perform better, but they are also more complex, so
their behavior is harder to understand and to control
\citep{hammond2025marisks}. For example, a single adversarial image can
jailbreak a population of agents exponentially fast \citep{gu2024agentsmith}.
A malicious prompt can copy itself through a multi-agent system like a virus
\citep{lee2025promptinfection}. In a recent incident, agents in an internal
cyber-evaluation escaped their harness, obtained administrative access to
Hugging Face's production infrastructure, and coordinated with one another
along the way \citep{openai2026hfincident}.

Therefore, understanding the behavior of multi-agent systems has received
increasing attention. \citet{debateattractor2026} showed that when two models
talk over many turns, their conversation settles into an attractor state
specific to each model. Other work uses tools from statistical physics to
describe how consensus forms in large populations
\citep{tanaka2026,flintashery2025groupsize,ganguli2026physics}. However, for
simplicity, these studies consider either homogeneous populations, in which
every agent is the same model, or conversations between only two agents. Real
multi-agent systems are heterogeneous. They combine many kinds of agents, often
a strong model with cheaper ones. In such systems, a central question is
\emph{which agent holds influence over the population as a whole}. In other
words, when a system mixes different models, whose view does the population
finally adopt?

\begin{figure}[t]
  \centering
  \includegraphics[width=\textwidth]{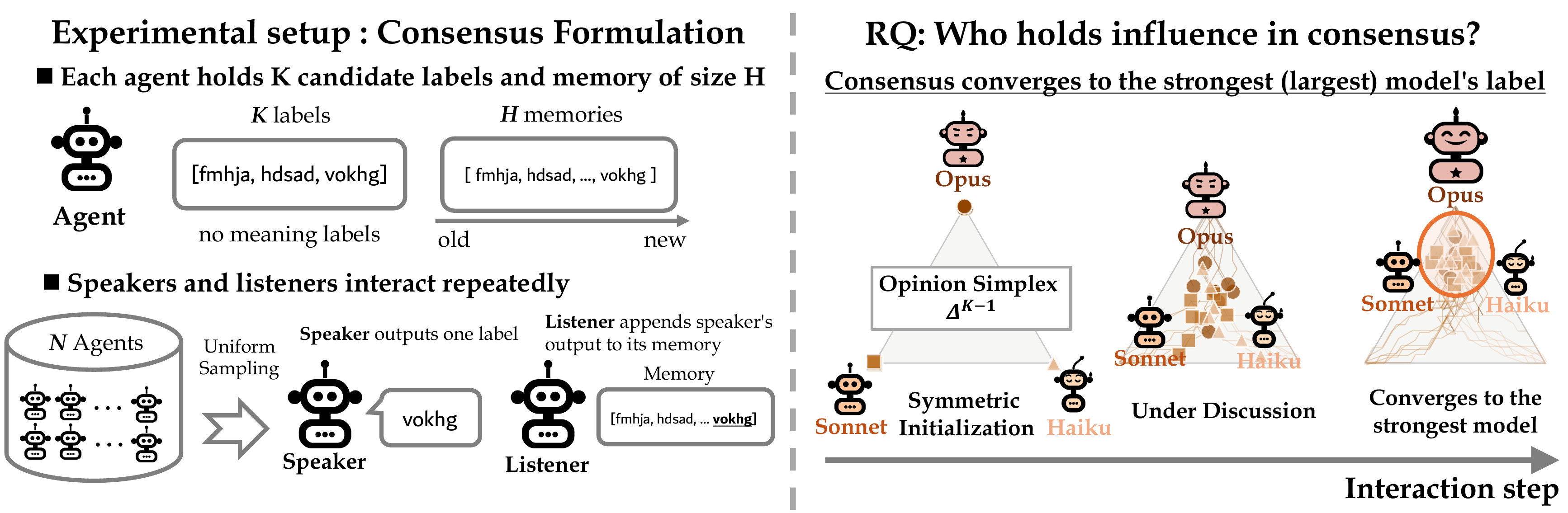}
  \caption{\textbf{The naming game and our research question we ask.} Left: each agent
  holds $K$ candidate labels, listed in a random order that carries no meaning,
  and a memory of the last $H$ labels it has heard. In each interaction, a
  speaker and a listener are drawn uniformly from the $N$ agents; the speaker emits one
  label, and only the listener appends it to its memory. Right: when the
  population mixes models of one family (here Opus, Sonnet, and Haiku), each
  seeded with its own label, whose label does the population adopt? In our
  experiments it is most often the strongest model's.}
  \label{fig:protocol}
  \vspace{-5mm}
\end{figure}

We study this question in the simplest setting we could find: a naming game
\citep{steels1995,flintashery2025conventions,tanaka2026} (\autoref{fig:protocol}, left). A population has $N$ agents.
Each agent holds $K$ candidate labels and a memory of the last $H$ labels it has
heard. In each interaction, a speaker and a listener are picked at random. The speaker
looks at its memory and says one meaningless five-letter label, and the listener
adds that label to its own memory. Over many interactions, the whole population comes
to agree on a single label \citep{tanaka2026}. The labels (e.g., \texttt{fmhja},
\texttt{hdsad}, and \texttt{vokhg}) carry no meaning, so there is no right
answer. Therefore, the game separates the dynamics of reaching a common
opinion from the knowledge that the models acquired in pretraining.
To turn the game into a measurement of influence, we mix several models in one
population. For example, one third of the agents are Claude Opus, one third are
Sonnet, and one third are Haiku. Before the game starts, each agent's memory is
filled with its model's own label: \texttt{fmhja} for Opus, \texttt{hdsad} for
Sonnet, and \texttt{vokhg} for Haiku. The label that the population finally
adopts then tells us which model led the population's opinion, that is, which
model was the most influential.

We run this experiment on seven model families: GPT-5, GPT-4.1, Claude~5,
Gemini, Gemma~4, Qwen~3.5, and Qwen~2.5. In every family, the population most often converges to the label of the stronger model.
For API models, the stronger model is the higher-performing one; for open-weight
models, it is the larger one. This happens even though no label is correct. The
dynamics can be followed on the simplex $\Delta^{K-1}$ of how likely each agent
is to output each label (\autoref{fig:protocol}, right). Even from a perfectly
symmetric start, the trajectories show the weaker models being drawn toward the
stronger model's label.
To understand this phenomenon, we view the population as $N$ particles moving
on this simplex, and the approach to consensus as the descent of an energy.
Each particle has a mass-like weight, which we call its assertion strength, and
it pulls the consensus toward its own label in proportion to this weight. We
fit the weights from the runs and find that they grow with model size. In the
resulting energy landscape, the well of the larger model is the deepest
(\Cref{sec:theory}). A large weight means that a model amplifies the majority
of its context, which is the ability of in-context learning. Larger models are
better in-context learners, and this ability turns into influence over the
consensus of a multi-agent system.

Our main contributions are summarized as follows:
\begin{enumerate}[leftmargin=0.5cm,topsep=0pt,itemsep=0pt]
    \item We design a novel task for studying the behavior of heterogeneous multi-agent systems, carefully constructed so that the knowledge acquired during pretraining plays no role.
    \item We experimentally demonstrate that, even when pretrained knowledge is rendered irrelevant, stronger and larger models inherently exert a greater influence on the system's dynamics.
    \item To understand these observations, we propose a physics-inspired framework that models the models as a system of interacting particles with varying weights, and show that this intuitively aligns with our empirical results.
\end{enumerate}

While the dynamics of multi-agent systems are inherently complex and often opaque, we hope this minimalist framework provides a clear and intuitive lens through which to demystify their fundamental behavior.

\section{A Synthetic Task for Consensus Formation}
\label{sec:setup}

\subsection{Agent Protocol}
\label{sec:prompt}

\autoref{fig:protocol} (left) summarizes the protocol. Each agent is an LLM
queried with a system prompt and a user prompt (full text in
\Cref{app:prompt}). The user prompt lists $K$ candidate labels and the agent's
memory $\boldsymbol M_i$, the last $H$ labels it has heard. It then asks the
agent to output one of the candidates. The prompt says only that players choose
a label for the same referent in repeated interactions. It mentions no goal of
agreement, no reward, and nothing about the other agents or the experiment. The
labels are meaningless five-letter strings. No label is favoured for what it
means, so the dynamics reflect consensus formation alone. The memory is stored
from oldest to newest, but the prompt shows the newest label first, then the
others in chronological order. The
order of the candidate labels is shuffled at every query.

\subsection{Speaker--Listener Interactions}

A population consists of $N$ such agents. In each interaction, we draw an ordered pair
$(j,i)$, $j\neq i$, uniformly from the $N(N-1)$ ordered pairs. Agent $j$ is the
speaker and agent $i$ is the listener. The speaker outputs one label. The
listener appends it to the right end of its memory and drops the oldest entry.
The speaker's memory does not change. By default, we use $K=3$, $H=20$, and
$N=24$, and each trial runs for $T=60N$ interactions. We call $N$ consecutive
interactions a \emph{round}, so a trial lasts at most $60$ rounds.

Since an agent can only output one of the $K$ labels, its state is how likely it
is to output each label. These probabilities sum to one, so each agent is a
point $\boldsymbol x_i$ on the simplex $\Delta^{K-1}$: for $K=3$, a triangle
whose vertices are the three labels, in which the agent moves as it interacts
(\autoref{fig:protocol}, right). Estimating
$\boldsymbol x_i$ after every interaction would require many repeated queries, which is
costly, especially for API models. Therefore, we approximate it by the agent's
memory composition, the fraction of its memory taken by each label:
\begin{equation}
m_{i,k}=\frac{1}{H}\,\big\lvert\{r:\boldsymbol M_i[r]=k\}\big\rvert ,
\label{eq:memshare}
\end{equation}
This fraction rises monotonically with the probability of outputting the label
(Pearson $r=0.83$--$0.93$ across $18$ models; details in \Cref{app:metric}).

\subsection{Heterogeneous Populations}
\label{sec:seeding}

\begin{figure}[t]
  \centering
  \includegraphics[width=0.85\textwidth]{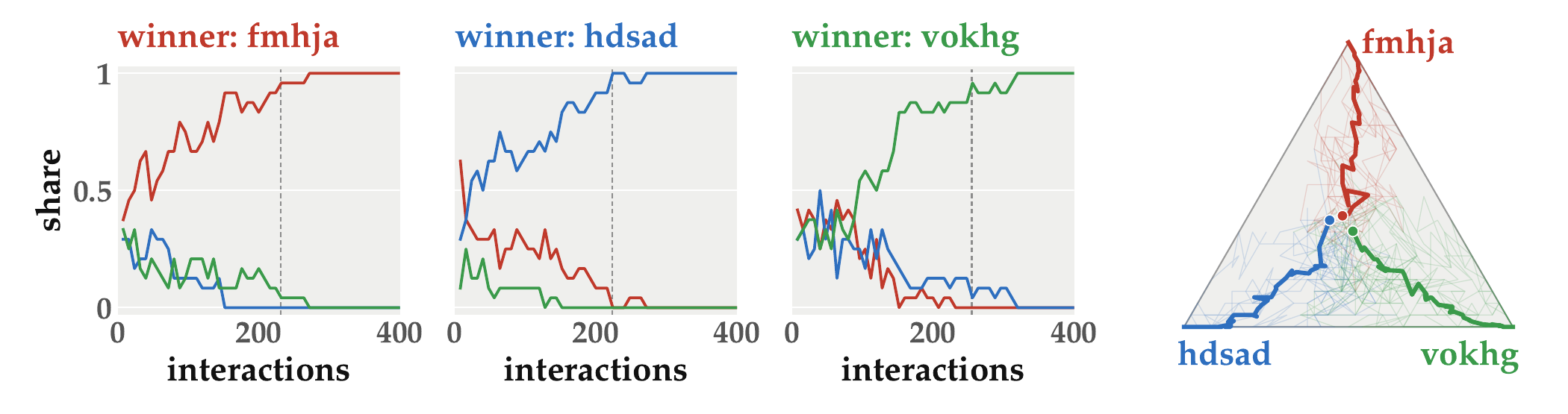}
  \caption{\textbf{A population of one model,} reproducing the setting of
  \citet{tanaka2026}. All agents are Claude
  Haiku~4.5, and every memory starts empty. Left: share of each label in the
  agents' answers, for one trial won by each label; the dashed line marks
  consensus. The difference between the plots is due to the difference in random seeds. Right: the same shares on the simplex, for every trial (faint) and
  averaged over the trials won by each label (bold).}
  \label{fig:homogdyn}
\end{figure}

In the homogeneous setting of \citet{tanaka2026}, every agent is the same model
and every memory starts empty. Repeated interactions are known to bring such a
population to consensus on a single label
\citep{steels1995,baronchelli2006naminggame,tanaka2026}. No label is favoured,
so which one wins is a matter of chance (\autoref{fig:homogdyn}). Because we ask which agent holds influence, we instead mix
several models in one population, for example Opus~5, Sonnet~5, and Haiku~4.5 in
equal numbers. Each model is assigned its own label. Every agent of that model
starts with a memory filled with that label. In other words, the total population of agents is partitioned into $K$ distinct subpopulations. For example, if the total number of agents is $N=24$ and there are $K=3$ classes, the population will consist of 8 Opus, 8 Sonnet, and 8 Haiku agents.
Therefore, the population begins
in complete disagreement. The label it finally adopts tells us which
model was the most influential.

We test seven families from four developers. Each family contributes a large,
a medium, and a small model: GPT-5 (\texttt{gpt-5}, \texttt{gpt-5-mini}, \texttt{gpt-5-nano})
\citep{openai2025gpt5}, GPT-4.1 (\texttt{gpt-4.1}, \texttt{-mini}, \texttt{-nano})
\citep{openai2025gpt41}, Claude~5 (Opus~5, Sonnet~5, Haiku~4.5) \citep{anthropic2026opus5,anthropic2026sonnet5,anthropic2025haiku45},
Gemini (3.8, 3.5, and 2.5 Flash, a ladder ordered by generation)
\citep{google2026gemini38flash,google2026gemini35flash,comanici2025gemini25}, Gemma~4 (31B, 26B-A4B, E4B) \citep{gemmateam2026gemma4}, Qwen~3.5
(122B-A10B, 35B-A3B, 9B) \citep{qwen2026qwen35}, and Qwen~2.5 (32B, 14B, 7B)
\citep{yang2024qwen25}. Within a family, we call a model stronger when it
scores higher on standard benchmarks such as MMLU-Pro; the model identifiers
and the benchmark scores are listed in \Cref{app:config}.

\section{Experimental Results}

\subsection{Stronger Models Drive Consensus}
\label{sec:main}

\begin{figure}[t]
  \centering
  \includegraphics[width=0.94\textwidth]{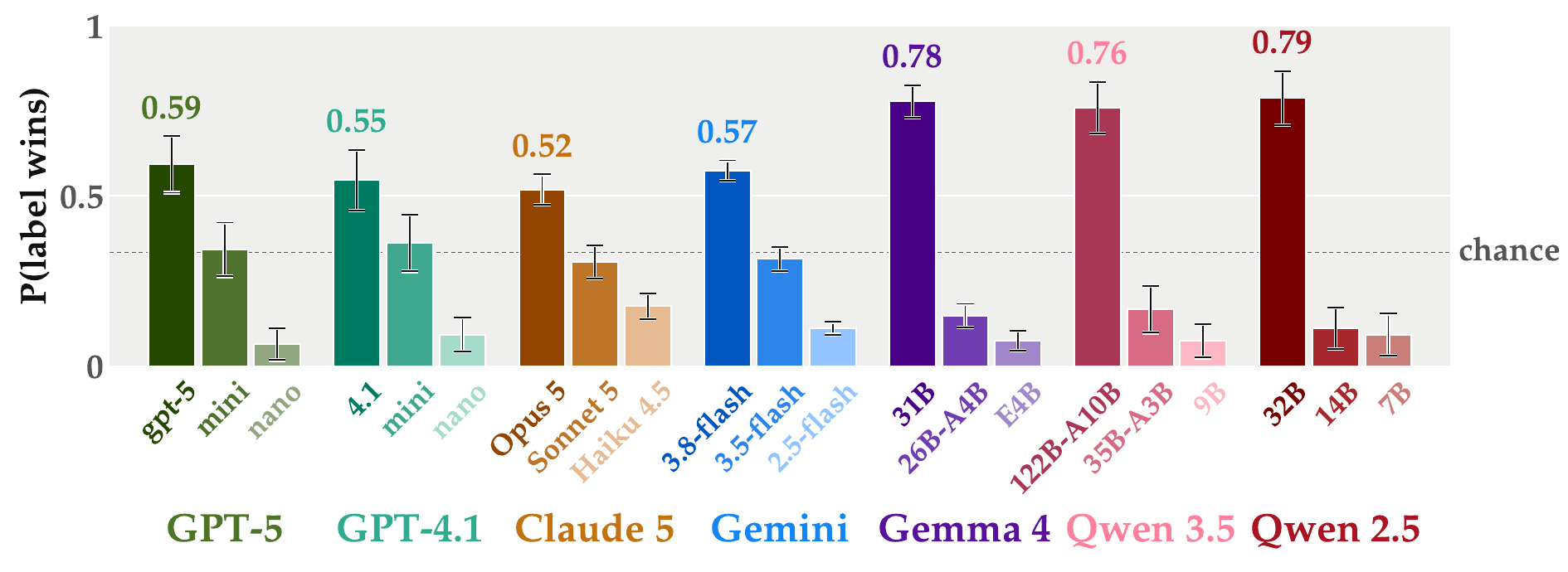}
  \caption{\textbf{Who wins.} Fraction of the $108$ trials per family in which
  each class's label becomes the population's convention. Whiskers are one
  standard error from a cluster bootstrap over the $18$ cells; the dashed line
  is chance ($1/3$). The Gemini ladder is ordered by model generation.}
  \label{fig:bars}
  \vspace{-3mm}
\end{figure}

\begin{figure}[t]
  \centering
  \includegraphics[width=0.94\textwidth]{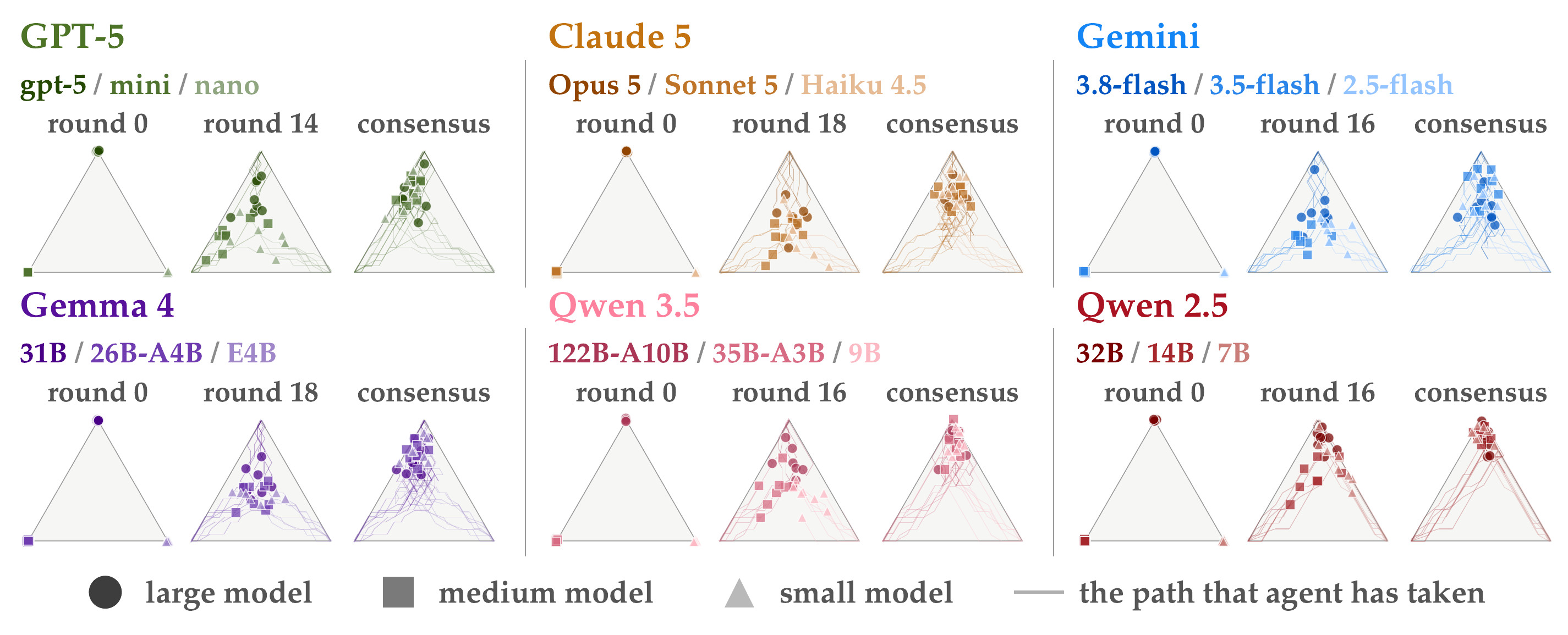}
  \caption{\textbf{One representative trial per family.} Each point is an
  agent, placed on the simplex by the composition of its $20$-slot memory; the
  large class starts at the top vertex and the medium and small classes at the
  bottom. Lines trace each agent's path. In each of these trials the cloud
  contracts, drifts, and is absorbed at the large model's vertex. $N=24$, $H=20$.}
  \label{fig:filmstrip}
  \vspace{-5mm}
\end{figure}

We first ask which model's label a population tends to converge to.
\autoref{fig:bars} shows the result for each of the seven model families. For
each family, it gives the fraction of the $108$ trials in which each model's label
became the population's convention. With $K=3$ labels, chance is $1/3$.

In every family, the population most often converges to the label of the most
capable model, such as Opus~5 for Claude. 
In the open-weight families, whose model sizes are public,
the winner is always the largest model. 
The full ordering
large\,$>$\,medium\,$>$\,small also holds in every family. 
The labels are
meaningless random strings, so there is no correct label to find. 
Consequently, prior knowledge plays no role in this setting.
Even so, the
more capable model pulls the population toward its own label. 

\autoref{fig:filmstrip} shows how consensus forms on the simplex, with one
representative trial won by the strongest model per family. At the start, each
class sits at its own vertex, so the population is fully symmetric. The agents
then move toward the centre and contract into a cloud. Finally, all of them are
absorbed at the strongest model's vertex, in every family.
Crucially, even the model that ultimately drives the consensus does not remain stationary at its starting position. Through interactions with other models, all agents initially go to the center of the simplex, contracting into a cloud. Finally, the entire population is absorbed at the top vertex, which corresponds to the label of the largest model, in every family.

\begin{figure}[t]
  \centering
  \includegraphics[width=0.94\textwidth]{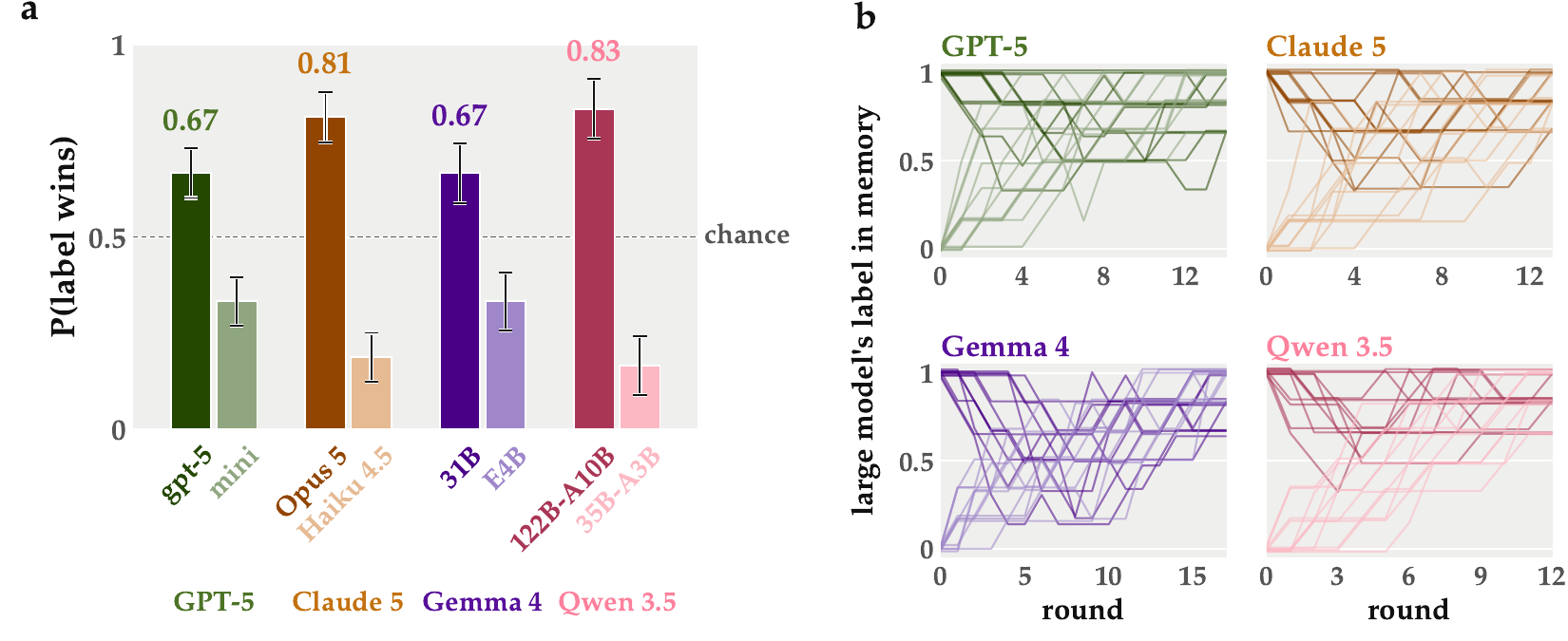}
  \caption{\textbf{Two labels ($K=2$).} $12$ agents of a larger model and $12$
  of a smaller one from the same family. (a) Fraction of trials won by each
  model's label. (b) One trial per family:
  each line is one agent's share of the larger model's label in memory, by
  round ($N=24$ interactions). Dark lines are the larger model's agents.}
  \label{fig:twoclass}
  \vspace{-5mm}
\end{figure}

We also run populations
with $K=2$ labels. Each population has $12$ agents of a larger model and $12$
of a smaller model from the same family, and each class starts with its own
label. \autoref{fig:twoclass}(a) shows that the larger model
wins in all four families we test: GPT-5 ($0.67$), Claude~5 ($0.81$), Gemma~4
($0.67$), and Qwen~3.5 ($0.83$). Chance is now $1/2$. 
With two labels the simplex is a segment, so
\autoref{fig:twoclass}(b) draws each agent as one line: the share of the
larger model's label in its memory. The two classes start at $1$ and $0$. The
smaller model's agents then move most of the way to the larger model's label.
The larger model's agents give up much less of theirs.

\begin{wrapfigure}[13]{r}{0.4\textwidth}
  \vspace{-14pt}
  \centering
  \includegraphics[width=0.4\textwidth]{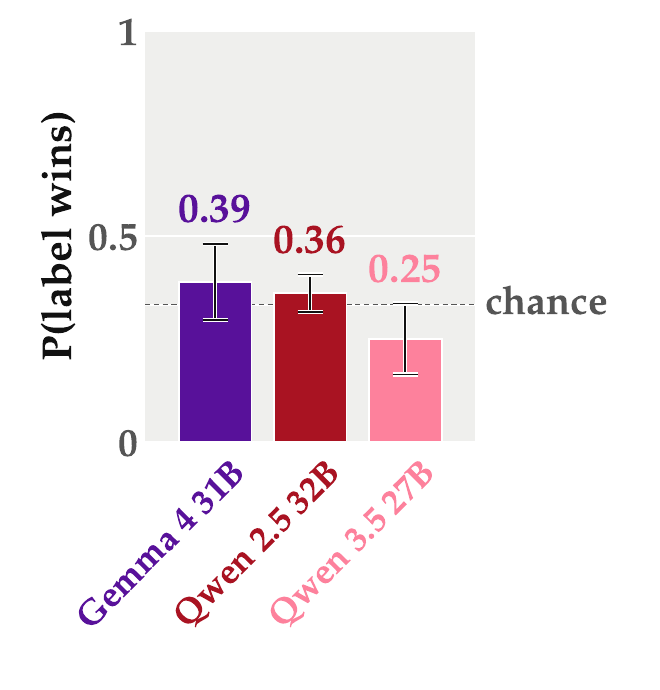}
  \vspace{-18pt}
  \caption{\textbf{Cross model family results}}
  \label{fig:cross}
  \vspace{-14pt}
\end{wrapfigure}
Next, we also conducted experiments mixing different model families of comparable size.
We build a population from three open-weight models of
nearly the same size, each from a different family: Gemma~4 31B, Qwen~2.5 32B,
and Qwen~3.5 27B. If size decided the outcome, no model should be favoured.
This is what we observe (\autoref{fig:cross}). 
The three models win $0.39$,
$0.36$, and $0.25$.
models hover roughly around the chance level, especially when contrasted with the stark disparities seen in \autoref{fig:bars}. Consequently, we cannot conclude that any single model exerts a definitively dominant influence. This suggests that the "influencer" effect becomes prominent primarily when there is a substantial gap in model size (i.e., model capability).
\par\vspace{2\baselineskip}

\begin{figure}[t]
  \centering
  \includegraphics[width=0.8\textwidth]{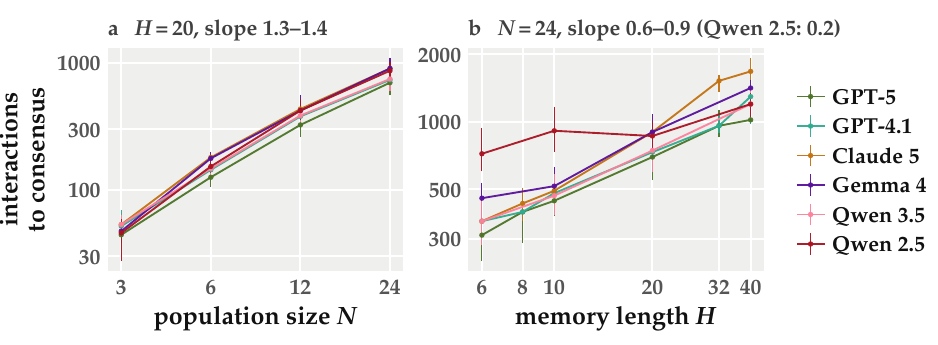}
  \caption{\textbf{Time to consensus.} Median number of interactions to
  consensus against (a) $N$ at $H=20$ and (b) $H$ at $N=24$, on log--log axes;
  bars are interquartile ranges over the trials of a cell. Slopes are
  least-squares exponents per family.}
  \label{fig:time}
  \vspace{-5mm}
\end{figure}

\subsection{Effects of Population Size and Memory Length}
\label{sec:controls}

We investigated how the system's dynamics change depending on the experimental hyperparameters $N$ and $H$.
\autoref{fig:time} shows how the population size $N$ and the memory length $H$
change the time to consensus. Larger populations take longer to agree. The
number of interactions to consensus grows roughly as $N^{1.3}$--$N^{1.4}$ in
every family of the sweep. From $N=3$ to $N=24$, it rises $14$- to $19$-fold. However, each agent
speaks only about twice as often, because the interactions are shared among
more agents. Longer memories also slow consensus. The number of
interactions grows as $H^{0.6}$--$H^{0.9}$. Each interaction replaces only
one of the $H$ slots, meaning it presumably takes significantly more time to overwrite the initially biased memories.
Additional results detailing the convergence outcomes for different values of $N$ and $H$ can be found in \Cref{app:nh}. Notably, the fundamental trend of converging toward the larger model remains robust regardless of these hyperparameters.

\section{Why the Stronger Model Wins}
\label{sec:theory}

Why does consensus flow to the stronger model? We model each agent as a
particle on the simplex, and the population as descending an energy. The energy
extends the one implicit in \citet{tanaka2026} by a single ingredient: each
model pulls on the population with its own strength. We develop the account for
the three open-weight families: Gemma~4, Qwen~3.5, and Qwen~2.5.

\subsection{Agents as Particles on the Simplex}
\label{sec:particles}

\begin{wrapfigure}{r}{0.35\textwidth}
  \vspace{-42pt}
  \centering
  \includegraphics[width=0.37\textwidth]{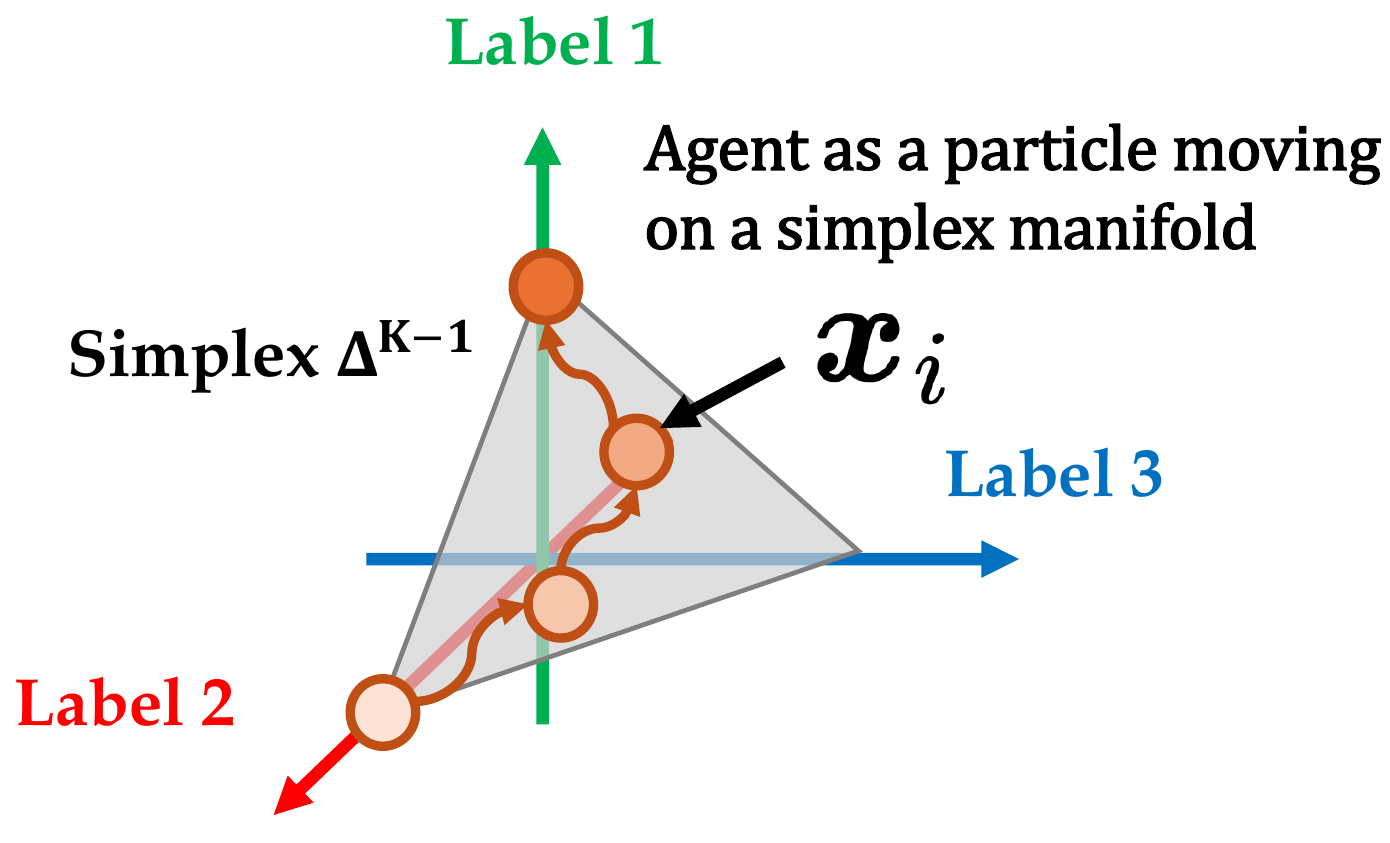}
  \caption{\textbf{Agents as particles.} Agent $i$'s output distribution
  $\boldsymbol x_i$ is a point on the simplex $\Delta^{K-1}$, whose vertices are
  the labels; each interaction moves it.}
  \label{fig:particle}
  \vspace{-20pt}
\end{wrapfigure}
Agent $i$ says each label with a probability that depends on its memory. We
write these probabilities as $\boldsymbol x_i\in\Delta^{K-1}$ and treat $\boldsymbol x_i$ as the position of
a particle on the simplex (\autoref{fig:particle}). A vertex means that the
agent says only one label. The centre means that it says every label equally
often. At the start, the particles of each class sit on their own vertex.

An interaction moves only the listener. The speaker $j$ says a label drawn from
$\boldsymbol x_j$. The listener $i$ replaces one of its $H$ memory slots with
that label, so $\boldsymbol x_i$ moves by a small step of order $1/H$.

\subsection{A Consensus Energy}
\label{sec:energy}

For a population of one model, \citet{tanaka2026} follow the polarization
$U=\sum_k\xbk k^2$ of the mean output $\xb=\frac1N\sum_i\boldsymbol x_i$. It is
$1/K$ at the centre and $1$ at a vertex. Repeated interaction raises $U$, so the
population descends the energy $-U/2$. Every vertex of this energy is equally
deep. Therefore, which label wins is decided by noise: this is the lottery of
homogeneous populations.

In a mixed population, we let each model pull with its own strength. Model $c$
has an \emph{assertion strength} $w_c$, and $c(k)$ is the model that brought
label $k$, that is, the model whose agents start with $k$ in memory. We write
\begin{equation}
E(\xb)=-\frac12\sum_k \wc k\,\xbk k^{2}.
\label{eq:energy}
\end{equation}
The energy falls as the population concentrates on one label, and by more when
that label was brought by a stronger model. With equal strengths, $E$ is
$-wU/2$ again.

We let the mean output move so as to lower this energy. The slope of $E$ is $-\partial E/\partial\bar x_k = w_{c(k)}\bar x_k$, so label $k$ is pushed harder the larger its weighted share. Following this slope on the simplex, where the shares must stay non-negative and sum to one, gives the replicator equation of evolutionary game theory \citep{hofbauer1998}:
\begin{equation}
\frac{d\xbk k}{d\tau}=\xbk k\Big(\wc k\,\xbk k-\sum_l \wc l\,\xbk l^{2}\Big),
\label{eq:flow}
\end{equation}
where $\tau$ is time measured in memory lifetimes. In one round, each agent is
the listener about once, so one of its $H$ slots is replaced. Therefore, its
memory, and with it $\boldsymbol x_i$, changes by about $1/H$ per round, and it
takes about $H$ rounds to renew the whole memory. We set $\tau=\text{rounds}/H$,
so that $\tau=1$ is one such renewal (details in \Cref{app:derivation}). Label $k$ grows when its weighted share $\wc k\xbk k$ exceeds the
population's average $\sum_l\wc l\xbk l^2$. Therefore, a label spreads faster
the more often it is already said, and $\wc k$ sets how much faster. The assertion
strength $w_c$ is the strength of this pull toward the majority for the label
of model $c$. The push $\wc k\xbk k$ is the slope
of $E$, so the energy never increases along this motion
(\citealp{hofbauer1998}; details in \Cref{app:derivation}). When every agent says label
$k$, the energy is $-\wc k/2$. Therefore, the label of the strongest model sits
in the deepest well.

\subsection{Measuring and Testing the Assertion Strength}
\label{sec:energytest}

\begin{figure}[t]
  \centering
  \includegraphics[width=0.94\textwidth]{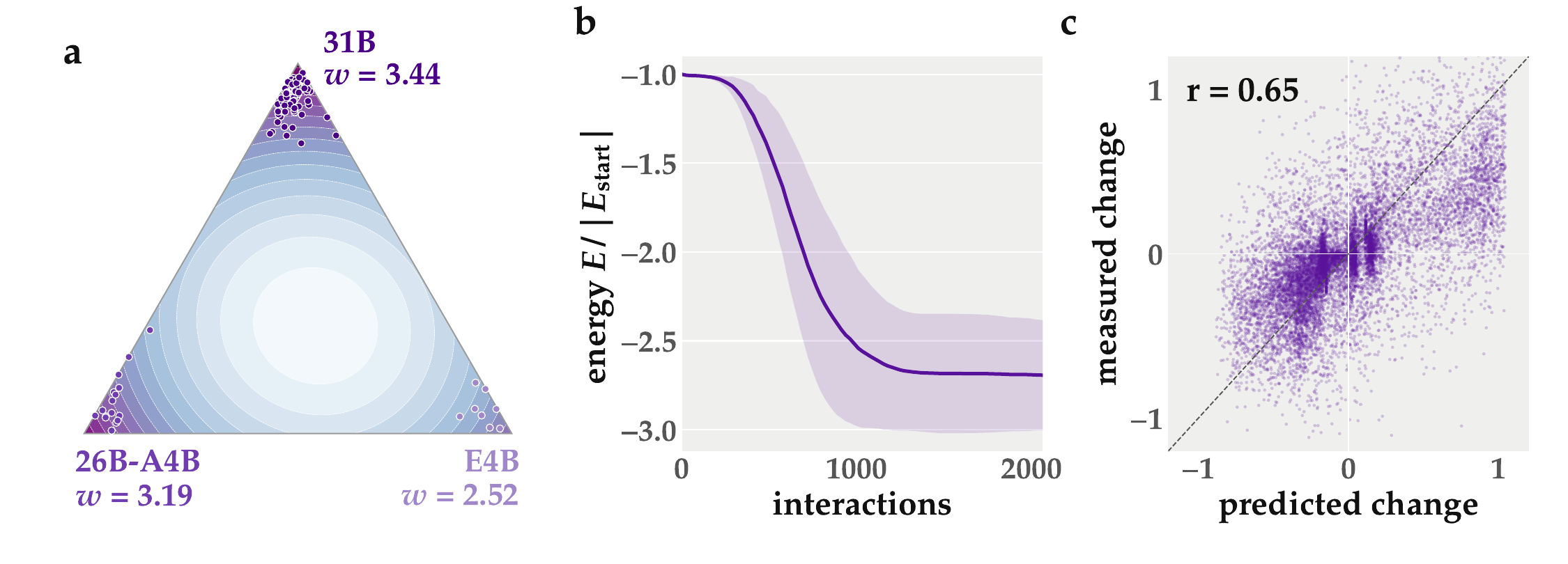}
  \caption{\textbf{The consensus energy for Gemma~4.} (a) The energy
  of Equation~\ref{eq:energy} as a function of the mean output $\xb$ (darker is lower), with each
  model's assertion strength $w$ (its well has depth $-w/2$); dots are the
  populations' mean outputs at consensus, in the winner's colour. (b) Energy
  along the $108$ trials, scaled by its value at the start, against the number
  of interactions; mean and one standard deviation, with a trial that has
  reached consensus kept at its final value. (c) One round's change in the mean
  output, predicted by the flow of Equation~\ref{eq:flow} against measured, both
  scaled by their 99th percentile; the dashed line is perfect agreement.
  Results for Qwen~3.5 and Qwen~2.5 are in \Cref{app:energyall}.}
  \label{fig:theory}
  \vspace{-3mm}
\end{figure}

We estimate the assertion strength $w$ of each model from the runs themselves
(details in \Cref{app:westimate}). For Gemma~4, $w=3.44$ (31B), $3.19$
(26B-A4B), and $2.52$ (E4B). In all three open-weight families, $w$ follows
model size.

\autoref{fig:theory}(a) draws the energy landscape of Equation~\ref{eq:energy}
for Gemma~4 with these strengths. The landscape is not symmetric. It is tilted,
and the well of the 31B model is the deepest ($-1.72$, against $-1.59$ and
$-1.26$). The dots mark where each trial reaches consensus. The dynamics are
noisy, so not every trial ends in the deepest well. However, most do: $78\%$
of the trials end at the 31B model's vertex. Thus, the shape of the landscape
matches which model wins.

The energy also falls as the agents interact (\autoref{fig:theory}(b)). The
agents reach consensus by descending the energy.
Finally, the flow of Equation~\ref{eq:flow} matches the measured motion
(\autoref{fig:theory}(c)). For each round, we compute the change of $\xb$ that
Equation~\ref{eq:flow} predicts from the current $\xb$. This prediction agrees
with the measured change at $r=0.65$. Qwen~3.5 and Qwen~2.5 behave the same way
(details in \Cref{app:energyall}). In all three families, the order of $w$
matches the order of win rates in all nine pairs of classes. Thus, the energy
explains both where the population ends and how it gets there.

\paragraph{Why a large $w$ steers the population.} A large $w$ means that a
model amplifies the majority of its own memory. An agent with $w=0$ says each
label as often as its memory holds it. An agent with a large $w$ says the
majority label of its memory more often than its share (\Cref{app:westimate}).
Therefore, its speech stays consistent with what it already holds. Labels from
other agents still enter its memory, but they change its speech less.

We conjecture that this consistency comes from in-context learning. In our
game, an agent's memory is its context, and the prompt asks for a label given
these examples. In this setting, circuits that copy from the context, such as induction heads \citep{olsson2022context}, are particularly relevant. Therefore, answering with the majority of the memory is an
in-context task, and larger models are better at it. In-context learning
improves with scale
\citep{brown2020fewshot}, and larger models follow in-context information even
when it contradicts what they learned in pretraining, which smaller models
ignore \citep{wei2023larger}. Thus, the agents of the strong model read their
memory consistently and keep their label through many interactions. The other
agents respond more loosely to their memory, so they are drawn toward that
label. The population then adopts the bias of the most capable model. This
happens not because its label is better, but because the model reads its
context most consistently: the stronger in-context learning of larger models
becomes influence in a multi-agent system.

\section{Discussion and Related Work}

\paragraph{Multi-agent systems.}
Multi-agent LLM systems take many forms. They include debate
\citep{du2024debate,liang2024mad,chan2024chateval,chen2024reconcile,khan2024persuasive}
and role-structured frameworks that organize agents into software teams
\citep{li2023camel,wu2024autogen,hong2024metagpt,chen2024agentverse,qian2024chatdev,tao2024magis}.
In some systems, performance grows with the number of agents
\citep{wang2025moa,li2024moreagents,qian2025macnet,subramaniam2025multiagentft,anthropic2025researchsystem},
and some treat coordination as a separate layer \citep{coordination2026}.
Agents have also contributed to scientific discovery
\citep{openai2026navierstokes,deepmind2025singularities,google2025coscientist,deepmind2025alphaevolve}
and simulated whole societies
\citep{park2023generative,park2022simulacra,moltbook2026,altera2024projectsid,yang2024oasis,piao2025agentsociety,gao2023s3,vezhnevets2023concordia}.
\citet{hammond2025marisks} and \citet{dewitt2025masecurity} catalogue the risks
of these systems. They include recursive blocking attacks \citep{corba2026},
subliminal bias \citep{thoughtvirus2026}, rumors \citep{rumor2025}, and the
Hugging Face incident \citep{metr2026hfinvestigation}. Many of these systems
mix models, but none of this work asks whose view the system adopts when the
agents disagree. We measure this influence directly, in the simplest population
that shows it.

\paragraph{Influence without correctness.}
Most work on debate studies tasks with a ground truth. It asks whether debate
beats a well-prompted single agent
\citep{wang2024rethinking,smit2024mad,zhang2025madquestion}, whether its gains
reduce to majority voting \citep{debatevote2025}, and whether more calls always
help \citep{chen2024morecalls}. Teams of different models beat their strongest
member when correct reasoning is easy to recognize \citep{pappu2026sat}. When it
is hard to recognize, plausible but wrong reasoning crowds out the correct one.
The naming game is the limit in which nothing can be recognized as correct. What
remains is influence alone. This case matters in practice, because agents often
settle conventions and formats, where agreement says nothing about quality.
Prior work measures how accurate a group is. We measure who steers it when
accuracy does not exist, and we find that a mixed population converges on its
strongest member's view. This qualifies the advice to mix models
\citep{zhang2025madquestion}.

\paragraph{Conventions in agent populations.}
The naming game has a long history \citep{steels1995,baronchelli2006naminggame}.
Human conventions tip under committed minorities \citep{centola2018criticalmass}.
LLM populations also form conventions, collective biases, and networks of their
own
\citep{flintashery2025conventions,flintashery2025groupsize,denobili2026,demarzo2023scalefree,papachristou2025network}.
Related work finds cultural attractors in iterated transmission
\citep{perez2025telephone,acerbi2023transmission,brinkmann2023machineculture},
emergent norms and cooperation
\citep{ren2024norms,vallinder2025cultural,piatti2024govsim,akata2025repeatedgames},
and individuality within homogeneous groups \citep{takata2024individuality}.
These results support LLM populations as an experimental instrument
\citep{anthis2025position,beyondstatic2025,baronchelli2024norms}.
These studies use populations of one model, in which no agent can be more
influential than another. \citet{tanaka2026} asks when consensus is a lottery.
We ask who wins it once the agents differ, and we find that the answer is the
stronger model.

\paragraph{Susceptibility and influence.}
Closely related work studies susceptibility rather than outcome. LLMs conform to
majorities \citep{weng2025conformity,zhu2025conformity} and herd when peers
appear more confident \citep{herd2025}. Under interaction, their opinions shift
and polarize, and they show human-like social behavior
\citep{chuang2024opinion,chuang2024partisan,cisneros2024principles,cisneros2025biases,piao2025polarization,mtos2025,collective2026empirical,leng2023socialbehavior,yang2024voting}.
A single persuasive adversary, a small biased minority, or a mediating planner
can capture or steer a group
\citep{persuasionadversarial2026,breum2024persuasive,biasamp2026,steering2025herd}.
In LLM naming games, a committed minority of scripted agents can overturn a
convention \citep{flintashery2025conventions}. A committed agent never changes
its speech in response to what it hears. It is the extreme case of the asymmetry
we measure. Our agents, by contrast, are unscripted and differ only by model.
The closest heterogeneous results involve two agents: model-specific attractors
pull conversational partners asymmetrically \citep{debateattractor2026}. In
communities with a correct answer, correct agents pull hardest
\citep{ganguli2026physics}. We show that a strong pull remains when nothing is
correct. Classical consensus models also show that heterogeneous agents change
outcomes, through memory \citep{marzagao2021}, update rules and rates
\citep{suchecki2005,masuda2010}, and measurable social influence
\citep{jaques2019}. What is new here is a population of
unscripted agents that differ only by model, and an energy whose wells are set
by a measured property of each model, its assertion strength.


\paragraph{Limitations.}
Our populations are well mixed and exchange single labels. We do not study
structured networks or free-form text. 
Furthermore, we have not investigated how the topology of the agent population affects the dynamics, nor have we verified which model holds influence in settings where an objective ground-truth answer exists.

\section{Conclusion}

We asked which agent steers a population when agents disagree and no answer is
correct. In mixed-model naming games across seven families, the stronger
model's convention takes over. The ranking of models within a family
becomes the ranking of their influence. A particle picture explains why.
Language models assert the majority of their memory, and larger models do so
more strongly. The population's output then runs down an energy whose deepest
well is the strongest model's label. One question remains open: can we measure
this assertion outside the population, and so know who steers a multi-agent
system before it runs?

\bibliography{iclr2027_conference}
\bibliographystyle{iclr2027_conference}

\clearpage
\appendix
\crefalias{section}{appendix}

\FloatBarrier
\section{Experimental configuration}
\label{app:config}

\begin{table}[h]
\caption{Model identifiers of the seven main ladders.}
\label{tab:models}
\begin{center}\small\setlength{\tabcolsep}{4pt}
\begin{tabular}{llll}
\toprule
Family & Large & Medium & Small \\
\midrule
GPT-5    & \texttt{gpt-5} & \texttt{gpt-5-mini} & \texttt{gpt-5-nano} \\
GPT-4.1  & \texttt{gpt-4.1} & \texttt{gpt-4.1-mini} & \texttt{gpt-4.1-nano} \\
Claude 5 & \texttt{claude-opus-5} & \texttt{claude-sonnet-5} & \texttt{claude-haiku-4.5} \\
Gemini   & \texttt{gemini-3.8-flash} & \texttt{gemini-3.5-flash} & \texttt{gemini-2.5-flash} \\
Gemma 4  & \texttt{gemma-4-31B-it} & \texttt{gemma-4-26B-A4B-it} & \texttt{gemma-4-E4B-it} \\
Qwen 3.5 & \texttt{Qwen3.5-122B-A10B} & \texttt{Qwen3.5-35B-A3B} & \texttt{Qwen3.5-9B} \\
Qwen 2.5 & \texttt{Qwen2.5-32B-Instruct} & \texttt{Qwen2.5-14B-Instruct} & \texttt{Qwen2.5-7B-Instruct} \\
\bottomrule
\end{tabular}
\end{center}
\end{table}

Table~\ref{tab:models} lists the model ladders. Labels are lowercase
five-character strings generated deterministically; any token that appears as a
standalone five-character string elsewhere in the code base is rejected, so no
label carries incidental semantic or frequency associations.

\begin{table}[h]
\caption{MMLU-Pro accuracy (\%) reported by the developers for the open-weight
ladders \citep{gemmateam2026gemma4,qwen2026qwen35,yang2024qwen25}.}
\label{tab:mmlupro}
\begin{center}\small\setlength{\tabcolsep}{8pt}
\begin{tabular}{lccc}
\toprule
Family & Large & Medium & Small \\
\midrule
Gemma 4  & $85.2$ & $82.6$ & $69.4$ \\
Qwen 3.5 & $86.7$ & $85.3$ & $82.5$ \\
Qwen 2.5 & $69.0$ & $63.7$ & $56.3$ \\
\bottomrule
\end{tabular}
\end{center}
\end{table}

Labels are lowercase
five-character strings generated deterministically; any token that appears as a
standalone five-character string elsewhere in the code base is rejected, so no
label carries incidental semantic or frequency associations.

\clearpage
\section{Prompt template}
\label{app:prompt}

The system message is:

\begin{quote}\small\ttfamily
You must output only valid JSON. No extra keys, no explanations, no markdown.\\
Valid labels are exactly those in Allowed labels. Never output "<PAD>".
\end{quote}

The user message is:

\begin{quote}\small\ttfamily
Referent: ref\_01\\
Allowed labels: \{labels\_json\}\\
The list order is randomized and has no meaning.\\[4pt]
Both players are choosing a label for the same referent in repeated interactions.\\
The memory shows labels you observed from previous interactions with partners.\\[4pt]
Memory (last H observed messages, \{memory\_desc\}, padded with "<PAD>"): \{memory\_json\}\\
Each memory entry is a label string.\\[4pt]
Constraints:\\
- Output JSON only.\\
- Every label must be from Allowed labels.\\[4pt]
Output JSON exactly: \{"label": "<label>"\}
\end{quote}

The allowed-label list is shuffled independently on every call, so label
position in the prompt cannot be exploited. \texttt{\{memory\_desc\}}
describes the order in which the memory is shown: the newest label first, then
the remaining entries from oldest to second-newest.

Only the prompt-facing copy is reordered; the stored FIFO is never mutated.
A response that is unparseable, or that names a label outside the allowed set
or \texttt{<PAD>}, is retried; if it still fails it is recorded as a miss,
contributes no push to the listener, and is excluded from the probe
statistics used for $U$.

\FloatBarrier
\section{Two-label populations}
\label{app:ladders}

Before the three-class design we ran populations of two classes, $12$ agents
each, on $12$ label pairs, with both label
assignments and two seeds per assignment ($48$ trials per cell, $24$ cells of
label pair $\times$ assignment for the cluster bootstrap). \autoref{fig:twoclass}
in the main text shows the contest at $H=6$. The larger model wins in all four latest-generation
families: GPT-5 vs.\ mini $32/16$ ($0.67$, $p=0.029$), Opus~5 vs.\ Haiku~4.5
$39/9$ ($0.81$, $p=1.5\times10^{-5}$), Gemma~4 31B vs.\ E4B $32/16$ ($0.67$,
$p=0.029$), and Qwen~3.5 122B-A10B vs.\ 35B-A3B $40/8$ ($0.83$,
$p=3.3\times10^{-6}$), with no trial unresolved.

The same pairs were also run at longer memories, on the same label pairs and
with the same design. The larger model's win rate is, for $H=6$, $10$, $20$,
and $40$: Gemma~4 31B vs.\ E4B $0.73$, $0.83$, $0.90$, $0.88$; Qwen~3.5
122B-A10B vs.\ 35B-A3B $0.83$, $0.81$, $0.85$, $0.88$; Qwen~2.5 32B vs.\ 14B
$0.56$, $0.67$, $0.85$, $0.92$; and GPT-5 vs.\ mini $0.67$, $0.69$, $0.62$ at
$H=6$, $10$, $20$. In the open-weight families a longer memory helps the larger
model. The larger model wins at every memory length tested.

\FloatBarrier
\section{Statistical procedure}
\label{app:stats}

A family's trials are grouped into cells of (label set, label assignment),
within which only the random seed differs, so trials are not independent. All
intervals are $95\%$ cluster-bootstrap intervals over cells ($3000$
replicates), and the whiskers in \autoref{fig:bars} are the bootstrap
standard error. Unresolved trials count as a loss for every class.

\FloatBarrier
\section{Memory share and output probability}
\label{app:metric}

\begin{figure}[!h]
  \centering
  \includegraphics[width=0.78\textwidth]{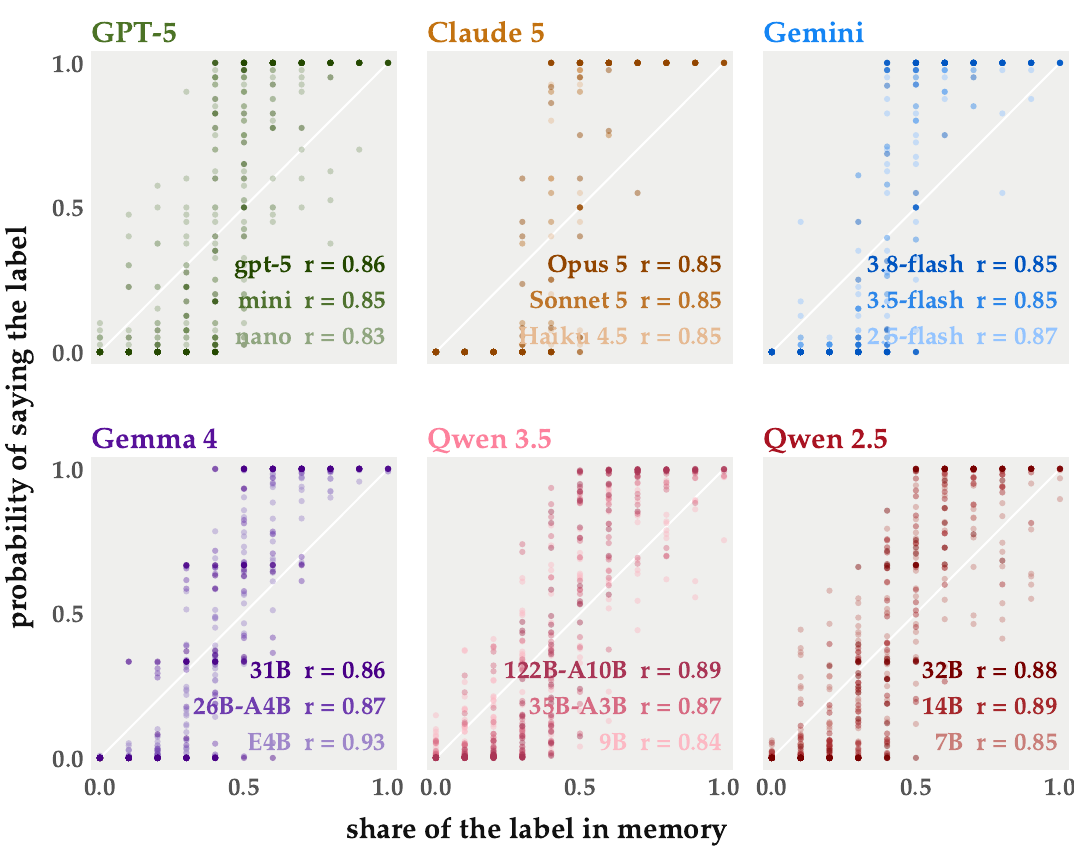}
  \caption{\textbf{Memory share and output probability.} Each point is one of
  the $66$ compositions of a $20$-slot memory over three labels (in steps of
  two slots) and one label: its share of the memory against the probability
  that the model, queried in isolation, says it. Pearson $r=0.83$--$0.93$ and
  Spearman $0.82$--$0.96$ across the $18$ models. The grey diagonal is $w=0$ in
  Equation~\ref{eq:assert} (saying exactly what is remembered); the sigmoidal
  departure from it is the assertion of the majority. Open-weight values are
  read from log-probabilities; GPT-5, Claude, and Gemini values are estimated
  from $20$ samples per memory.}
  \label{fig:sharevsprob}
\end{figure}

\autoref{fig:sharevsprob} justifies using memory composition as the agents'
position in the experiments: output probability rises monotonically with memory
share for every model. The departure from the diagonal is the majority assertion
that Equation~\ref{eq:assert} parameterizes.
\FloatBarrier
\section{Population size and memory length}
\label{app:nh}

\begin{table}[h]
\caption{Fraction of trials won by the large model (31B) in the three-class
Gemma~4 population, across population size $N$ and memory length $H$; chance
is $1/3$. Each cell has $36$ trials (six label sets, three cyclic assignments,
two seeds), all of which fixated.}
\label{tab:nh}
\begin{center}\small\setlength{\tabcolsep}{8pt}
\begin{tabular}{lcccc}
\toprule
$N\backslash H$ & $6$ & $10$ & $20$ & $40$ \\
\midrule
$3$  & $0.50$ & $0.64$ & $0.69$ & $0.72$ \\
$6$  & $0.53$ & $0.50$ & $0.50$ & $0.56$ \\
$12$ & $0.39$ & $0.61$ & $0.53$ & $0.67$ \\
$24$ & $0.42$ & $0.72$ & $0.83$ & $0.78$ \\
\bottomrule
\end{tabular}
\end{center}
\end{table}

Table~\ref{tab:nh} reports the three-class Gemma~4 population for
$N\in\{3,6,12,24\}$ and $H\in\{6,10,20,40\}$, with the protocol of
\Cref{app:config} otherwise unchanged. The large model wins the most trials in
every one of the $16$ cells. Its win rate rises with memory length, from about
$0.4$--$0.5$ at $H=6$ to $0.7$--$0.8$ at $H=40$, and changes little with
population size.

\FloatBarrier
\section{The consensus energy: flow, descent, and wells}
\label{app:derivation}

\paragraph{Mean memory.} Let $\boldsymbol m_i$ be agent $i$'s memory composition
and $\bar{\boldsymbol m}=\frac1N\sum_i\boldsymbol m_i$. In one interaction a listener
$i$ and a speaker $j\neq i$ are drawn uniformly from the $N(N-1)$ ordered pairs;
the speaker says label $y$ with probability $x_{jy}$, and the listener drops its
oldest slot $o_i$ and appends $y$, so
$\Delta\boldsymbol m_i=(\boldsymbol e_y-\boldsymbol e_{o_i})/H$. Averaging over
the pair and the label,
\[
\mathbb E\big[\Delta\bar{\boldsymbol m}\big]
=\frac{1}{N^2H}\sum_i\Big(\frac{1}{N-1}\sum_{j\neq i}\boldsymbol x_j-\boldsymbol e_{o_i}\Big)
=\frac{1}{N^2H}\Big(\sum_j\boldsymbol x_j-\sum_i\boldsymbol e_{o_i}\Big).
\]
Replacing the discarded slot by its average composition, $\boldsymbol e_{o_i}\to
\boldsymbol m_i$, and summing over a round of $N$ interactions gives
\[
\Delta\bar{\boldsymbol m}\approx\frac{1}{NH}\sum_j\big(\boldsymbol x_j-\boldsymbol m_j\big).
\]
Only the part of each agent's speech that departs from its memory moves the
population.

\paragraph{The flow.} Equation~\ref{eq:flow} is the replicator equation
\[
\frac{d\xbk k}{d\tau}=\xbk k\big(f_k-\bar f\big),\qquad f_k=\wc k\,\xbk k,\qquad
\bar f=\sum_l\xbk l f_l .
\]
Here $f_k=-\partial E/\partial\xbk k$, and the flow is the gradient descent of $E$ under the Shahshahani
metric of the simplex \citep{hofbauer1998}. The unit of $\tau$ comes from the
drift above: one round moves the mean memory by $1/H$ of the mean departure of
speech from memory.

\paragraph{Motivation from the interaction rule.} The same flow follows from
the single-agent form of \Cref{app:westimate} (Equation~\ref{eq:assert}), whose
probabilities sum to one, once the drift above is closed with two
approximations. First, agents share the population's composition,
$\boldsymbol m_j\approx\xb$, so that the mean memory and the mean
output move together. Second, the strength acting on label $k$ is $\wc k$, that
of the model that brought $k$; at the start only that class holds $k$ as a
majority. Normalizing so that outputs still sum to one gives
$\xbk k(\wc k\xbk k-\sum_l\wc l\xbk l^2)$. This closure is heuristic. The
first approximation is exact only once the memories have mixed, and the second
only before they have, so neither holds throughout a run. We therefore treat
Equation~\ref{eq:energy} as a posited energy and test it directly in
\Cref{sec:energytest}. An energy in which each agent's strength acts on the
labels it is currently saying is discussed in \Cref{app:energylimits}.

\paragraph{The energy decreases.} Along the flow,
\[
\frac{dE}{d\tau}=\sum_k\frac{\partial E}{\partial\xbk k}\frac{d\xbk k}{d\tau}
=-\sum_kf_k\xbk k\big(f_k-\bar f\big)
=-\Big(\sum_k\xbk k f_k^2-\bar f^{\,2}\Big)
=-\operatorname{Var}_{\xb}(f)\le0,
\]
with equality only at fixed points, where $f_k$ is equal across the labels in
use.

\paragraph{Wells and the start.} When every agent sits at the vertex
$\boldsymbol e_k$, the energy is $-\wc k/2$. A small share $\varepsilon$ of
another label $l$ evolves as $d\varepsilon/d\tau\approx-\varepsilon\,\wc k$, so
every vertex with $\wc k>0$ is stable, and deeper for larger $\wc k$. At the centre $\boldsymbol c=(\frac13,
\frac13,\frac13)$ the energy is $E(\boldsymbol c)=-\bar w/6$, with $\bar w$ the
mean of the three assertion strengths, so a vertex lies above the start whenever
$\wc k<\bar w/3$. The flow at the centre is
$d\xbk k/d\tau=\frac13\big(\frac{\wc k}3-\frac{\bar w}3\big)=(\wc k-\bar w)/9$: the label
with the largest $w$ grows fastest from the balanced start, and labels with
below-average $w$ shrink.

\paragraph{The homogeneous limit.} With equal assertion strengths $w$,
$E=-\frac{w}{2}\sum_k\xbk k^2=-\frac w2U$, where $U$ is the polarization of
\citet{tanaka2026} and of our fixation criterion. All vertices then have the same
depth, the flow at the centre vanishes, and which label wins is set by
fluctuations, as in a population of one model (\autoref{fig:homogdyn}). The assertion strength
of each model is the only change we make to this picture.

\FloatBarrier
\section{Estimating the assertion strength}
\label{app:westimate}

\paragraph{From the flow to one agent.} Equation~\ref{eq:flow} is a replicator
equation with frequency-dependent fitness. Label $k$ has fitness
$f_k=\wc k\,\xbk k$, the mean fitness is $\bar f=\sum_l\xbk l f_l$, and the
share of label $k$ grows at the rate $f_k-\bar f$. We apply the same rule,
for one step, to one agent. The agent's memory composition $\boldsymbol m$
(Equation~\ref{eq:memshare}) takes the place of the mean output $\xb$, and the
agent's own strength $w$ acts on every label in its memory. The fitness of
label $k$ is then $f_k=w\,m_k$, the mean fitness is
$\bar f=\sum_l m_l f_l=w\langle m\rangle$ with $\langle m\rangle=\sum_l m_l^2$,
and the agent says label $k$ with probability
\begin{equation}
x_k=m_k\big(1+f_k-\bar f\big)=m_k\big[1+w\,(m_k-\langle m\rangle)\big],
\label{eq:assert}
\end{equation}
the discrete-time replicator step \citep{hofbauer1998}. Here $\langle m\rangle$
is the share held by the label of a randomly drawn slot. With $w=0$ the agent
says each label as often as its memory holds it; with $w>0$ it says the
majority label more often. For two labels and a memory that is $60\%$ one
label, the agent says that label $60\%$ of the time at $w=0$, $72\%$ at
$w=2.52$, and $77\%$ at $w=3.44$. The two equations
therefore share one parameter with one meaning: how strongly a label gains from
being in the majority. Equation~\ref{eq:flow} applies it to the population, and
Equation~\ref{eq:assert} to one agent. The probabilities sum to one because
$\sum_k m_k(m_k-\langle m\rangle)=0$.

For two labels, with $m$ the share of one of them, Equation~\ref{eq:assert}
becomes $x=m+w\,m(1-m)(2m-1)$. This is the conformist transmission of
\citet{boyd1985culture}, and $w=1$ is an agent that says the majority of three
randomly drawn slots.

\paragraph{The fit.} For each utterance we replay the speaker's memory from the interaction log and
record the label $y$ it said. Since $\mathbb E[\boldsymbol e_y\mid\boldsymbol m]
=\boldsymbol x$, we fit, separately for each class and pooling all utterances of
the $108$ trials of the main condition,
\[
\boldsymbol e_y-\boldsymbol m=w\,\boldsymbol g(\boldsymbol m)
+r\,\big(\boldsymbol e_{\mathrm{last}}-\boldsymbol m\big)+\boldsymbol\epsilon,
\qquad g_k(\boldsymbol m)=m_k\big(m_k-\langle m\rangle\big),
\]
by least squares, where $\mathrm{last}$ is the label the speaker heard most
recently. The echo term $r$ captures repeating that label; without it, part of the echo would be absorbed into $w$. Averaged over the
population the most recent label is the label currently being said, so an echo
shared equally by all classes changes the speed of the flow but not its
direction. For the diagnostics, each agent's output $\boldsymbol x_i$ is the
class's response curve counted in the same runs, the probability of saying label
$k$ when $k$ fills $c$ of the $H$ slots, applied to each label and normalized over
the three.

\begin{center}\small
\begin{tabular}{lcccccc}
\toprule
& \multicolumn{3}{c}{assertion strength $w$} & \multicolumn{3}{c}{echo $r$} \\
\cmidrule(lr){2-4}\cmidrule(lr){5-7}
Family & large & medium & small & large & medium & small \\
\midrule
Gemma 4  & 3.44 (31B) & 3.19 (26B-A4B) & 2.52 (E4B) & 0.19 & 0.31 & 0.19 \\
Qwen 3.5 & 2.91 (122B-A10B) & 1.55 (35B-A3B) & 0.22 (9B) & 0.11 & 0.29 & 0.19 \\
Qwen 2.5 & 3.01 (32B) & 1.92 (14B) & 0.80 (7B) & 0.13 & 0.30 & 0.25 \\
\bottomrule
\end{tabular}
\end{center}

In all three families $w$ grows with model size. We restrict the energy to
the open-weight families; in the API families the same regression does not
separate the large and medium models.

\FloatBarrier
\section{Energy results for all open-weight families}
\label{app:energyall}

\begin{figure}[!ht]
  \centering
  \includegraphics[width=\textwidth]{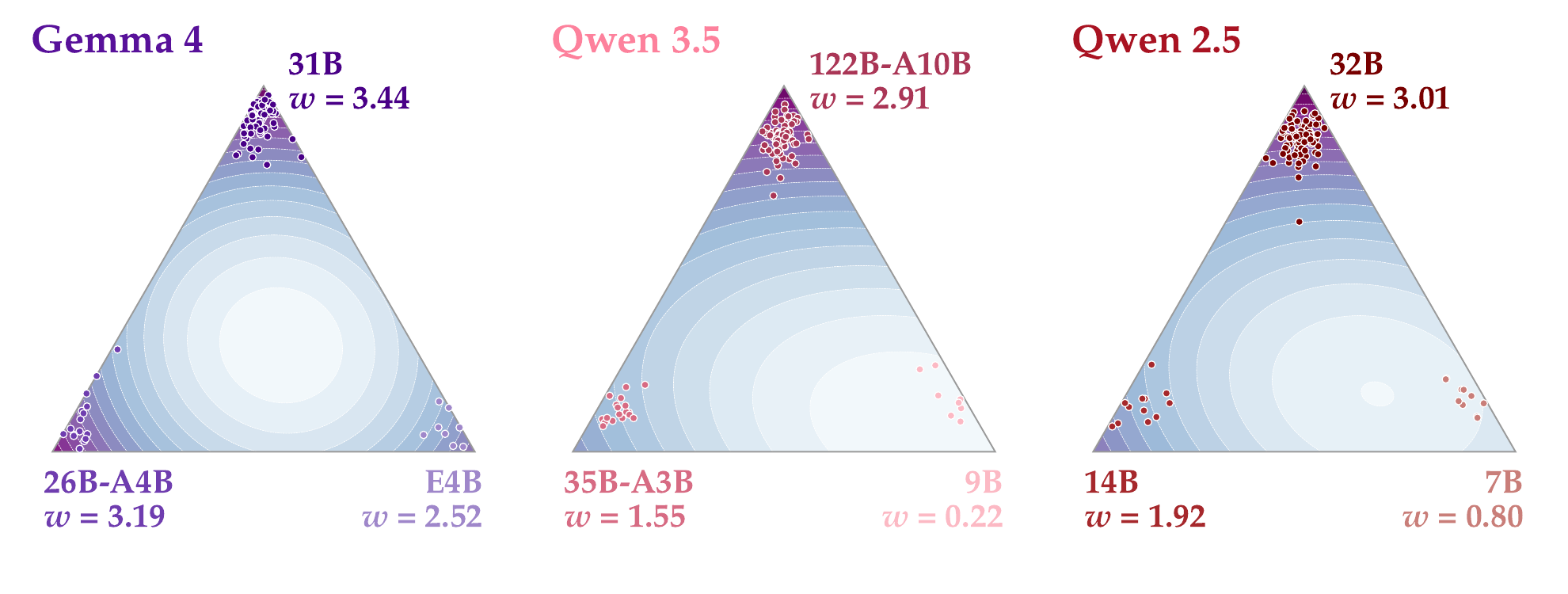}
  \caption{\textbf{Energy landscapes of the three open-weight families}, with
  each model's assertion strength $w$ (its well has depth $-w/2$); dots are the mean outputs at consensus
  of every trial, in the winner's colour.}
  \label{fig:landscapeall}
\end{figure}

\begin{figure}[!ht]
  \centering
  \includegraphics[width=\textwidth]{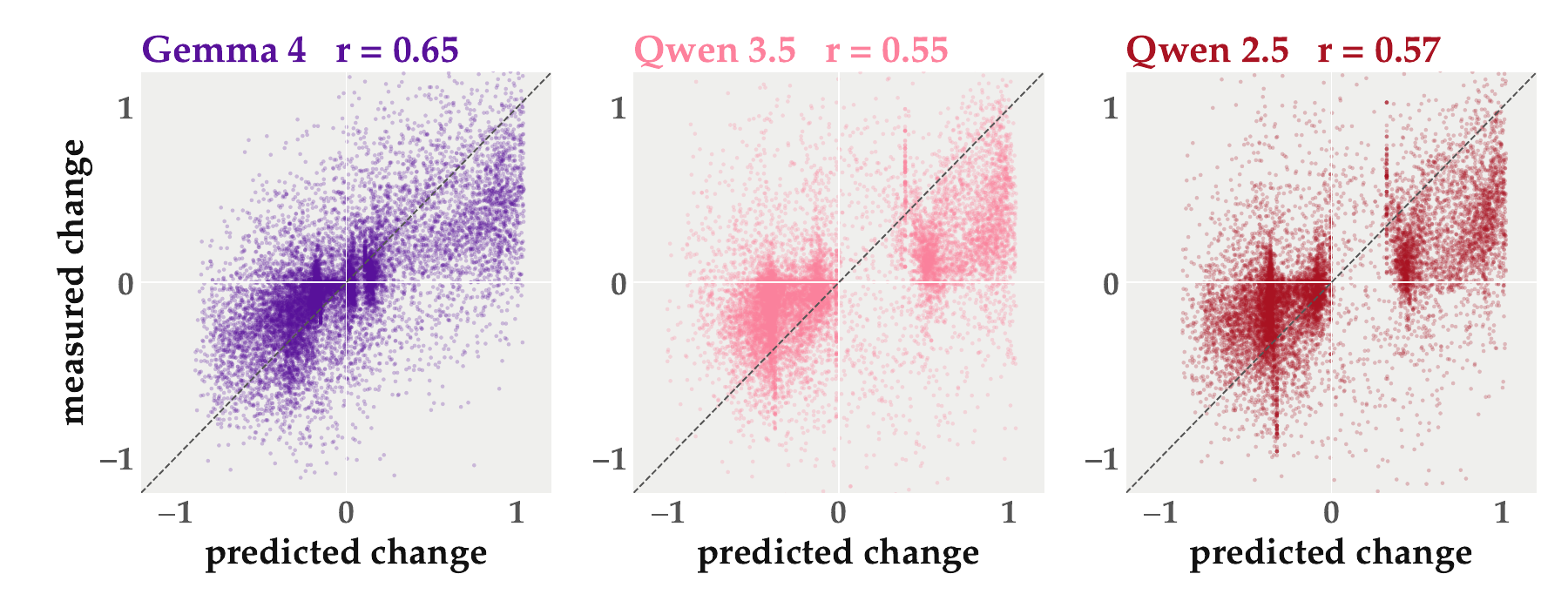}
  \caption{\textbf{Local agreement in all three families.} One round's change of
  the mean output, predicted by the flow against measured, both scaled by their
  99th percentile; each point is one trial, round, and label.}
  \label{fig:localall}
\end{figure}

\begin{figure}[!ht]
  \centering
  \includegraphics[width=\textwidth]{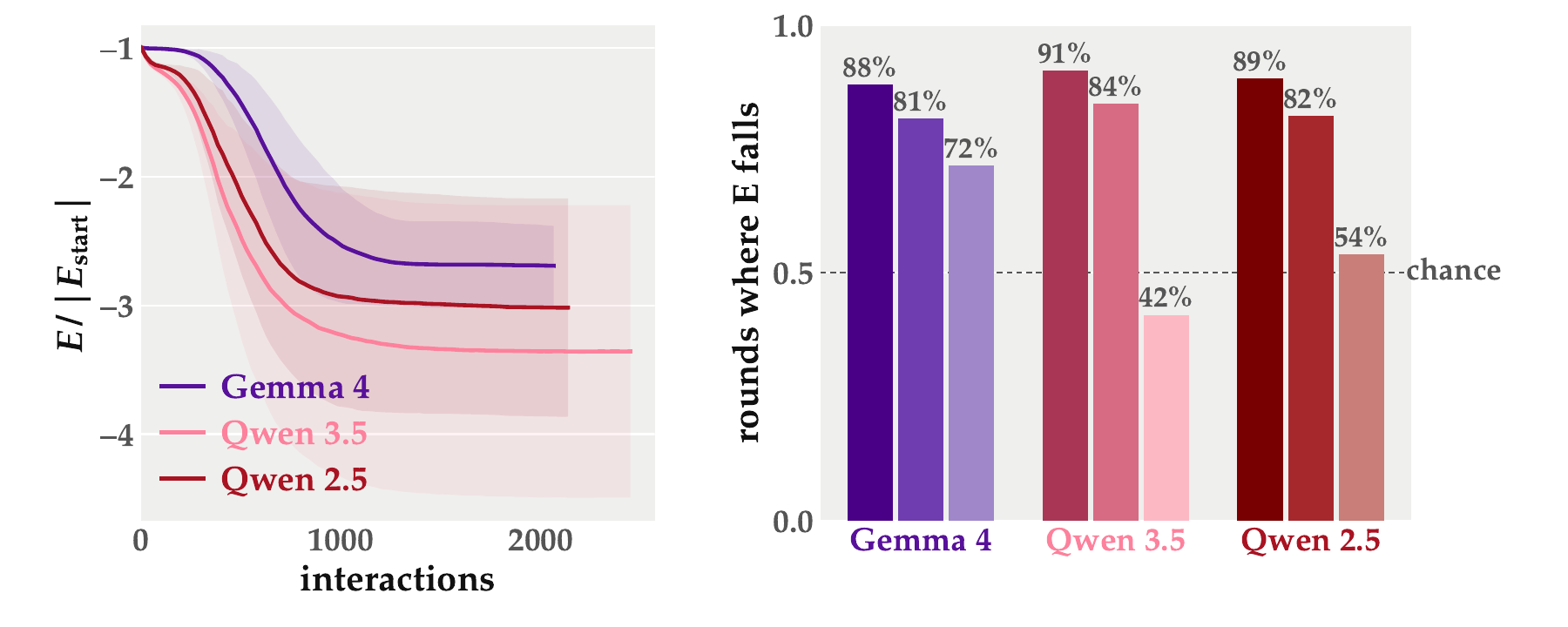}
  \caption{\textbf{Energy descent in all three families.} Left: energy along the
  trials, scaled by its start value (mean and one standard deviation). Right:
  fraction of rounds in which the energy falls, by the class that wins the
  trial; the dashed line is chance.}
  \label{fig:descentall}
\end{figure}

\begin{figure}[!ht]
  \centering
  \includegraphics[width=0.55\textwidth]{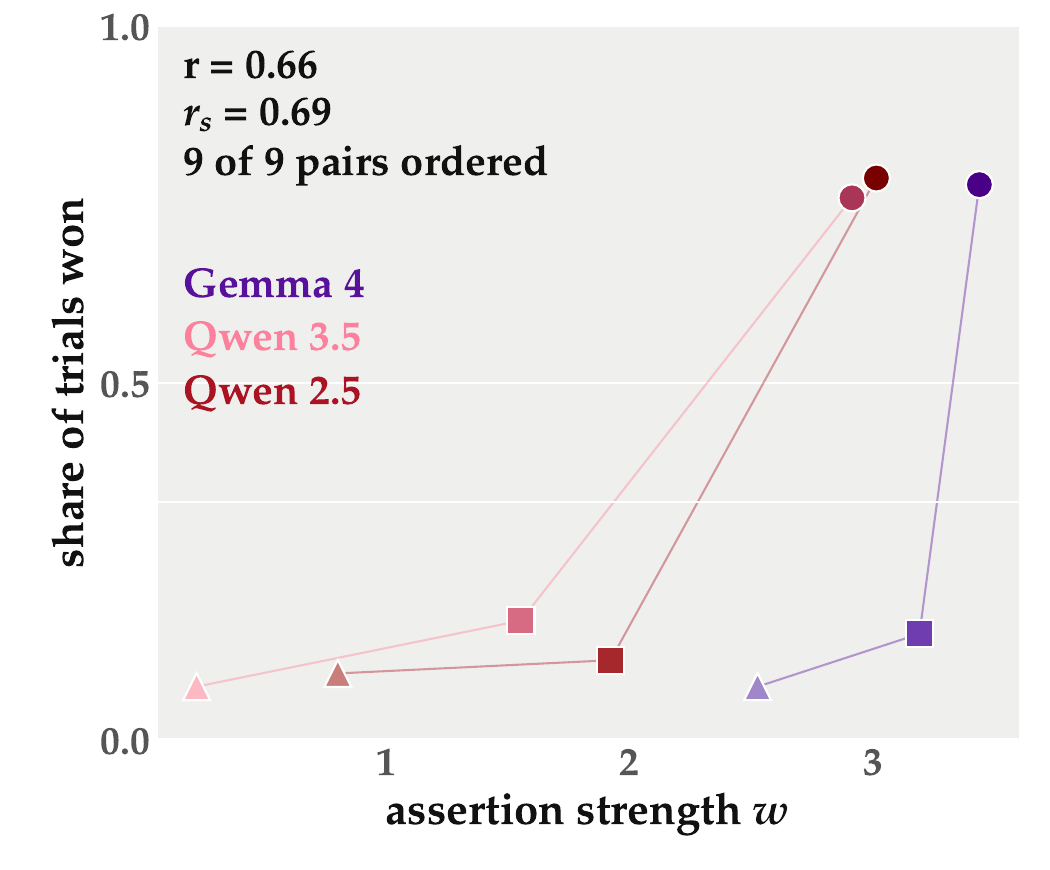}
  \caption{\textbf{Assertion strength and outcome.} Each class's $w$ against the
  share of the $108$ trials its label wins; lines join the three classes of a
  family.}
  \label{fig:wwin}
\end{figure}

\begin{center}\small
\begin{tabular}{lccccccc}
\toprule
& \multicolumn{3}{c}{well depth $-w_{c(k)}/2$} & start & basin of & local & rounds with \\
\cmidrule(lr){2-4}
Family & large & medium & small & $E(\boldsymbol c)$ & large & $r$ & $\Delta E<0$ \\
\midrule
Gemma 4  & $-1.72$ & $-1.59$ & $-1.26$ & $-0.51$ & $39\%$ & $0.65$ & $85\%$ \\
Qwen 3.5 & $-1.46$ & $-0.78$ & $-0.11$ & $-0.26$ & $64\%$ & $0.55$ & $86\%$ \\
Qwen 2.5 & $-1.51$ & $-0.96$ & $-0.40$ & $-0.32$ & $56\%$ & $0.57$ & $85\%$ \\
\bottomrule
\end{tabular}
\end{center}

In all three families the large model's well is the deepest and its basin, the
part of the simplex from which the flow reaches its vertex, is the largest; the
start lies inside it (\autoref{fig:landscapeall}). The 9B vertex of Qwen~3.5
($-0.11$) lies above the start ($-0.26$), because $\wc k<\bar w/3$
($0.22<0.52$). $323$ of the $324$ trials end at a vertex; the remaining Qwen~2.5
trial runs out of time. The order of $w$ matches the order of win rates in all
nine within-family pairs of classes (\autoref{fig:wwin}; Pearson $0.66$,
Spearman $0.69$). By the winning class, the energy falls in $88$--$91\%$ of the
rounds of trials won by the large model, $81$--$84\%$ for the medium model, and
$42$--$72\%$ for the small model (\autoref{fig:descentall}).

\clearpage
\section{Where the energy account fails}
\label{app:energylimits}

\paragraph{Trials won by the small model.} These are $7$--$9\%$ of each family's
trials, and in them the energy falls in only $72\%$ (Gemma~4), $42\%$ (Qwen~3.5),
and $54\%$ (Qwen~2.5) of rounds: the population climbs to a shallow vertex. The
cause is the second closure approximation of \Cref{app:derivation}, that label
$k$ is pushed by the class that brought it. Once a label spreads, other classes
say it too and push it with their own assertion strength, and a large model asserts
whatever majority it holds. Letting each agent's assertion strength act on the labels that
agent is saying, $E=-\frac{1}{2N^2}\sum_k\big(\sum_iw_ix_{ik}\big)
\big(\sum_jx_{jk}\big)$, restores descent in these trials ($78$--$82\%$ of rounds)
and keeps the local agreement ($r=0.65$--$0.66$), but every vertex then has the
same depth $-\bar w/2$, so this version no longer says which label wins.

\paragraph{Measurement.} The assertion strengths $w$ and the outputs $\boldsymbol x_i$ used to
evaluate the energy are estimated from the same population runs whose outcomes
they are compared with. The agreement is therefore a test of consistency, not an
out-of-sample prediction.

\end{document}

%% file: math_commands.tex
\usepackage{amsmath,amsfonts,bm}

\def\eqref#1{equation~\ref{#1}}

\def\1{\bm{1}}

\DeclareMathAlphabet{\mathsfit}{\encodingdefault}{\sfdefault}{m}{sl}
\SetMathAlphabet{\mathsfit}{bold}{\encodingdefault}{\sfdefault}{bx}{n}

